\documentclass[letterpaper, 10 pt, conference]{ieeeconf}  % Comment this line out if you need a4paper

\IEEEoverridecommandlockouts                              % This command is only needed if 
\usepackage{graphicx}
\usepackage{amssymb}
\usepackage{xcolor}
\usepackage{amsmath}
\usepackage{flushend} % aligned references columns
\usepackage{hyperref}
\title{\LARGE \bf
Boosting Reinforcement Learning Algorithms in Continuous Robotic Reaching Tasks using Adaptive Potential Functions*
}

\author{Yifei Chen$^{1}$, Lambert Schomaker$^{1}$, and Francisco Cruz$^{2,3}$% <-this % stops a space
\thanks{*This work was conducted during Yifei Chen's research stay at the robotics lab of the University of New South Wales.}% <-this % stops a space
\thanks{$^{1}$Yifei Chen and Lambert Schomaker are with Bernoulli Institute for Mathematics, Computer Science and Artificial Intelligence, University of Groningen, The Netherlands
        {\tt\small \{yifei.chen,l.r.b.schomaker\}@rug.nl}}%
\thanks{$^{2}$Francisco Cruz is with School of Computer Science and Engineering, University of
New South Wales, Sydney, NSW, Australia
        {\tt\small f.cruz@unsw.edu.au}}%
\thanks{$^{3}$Francisco Cruz is with Escuela de Ingenieria, Universidad Central de Chile,
Santiago, Chile}%
}

\begin{document}

\maketitle
\thispagestyle{empty}
\pagestyle{empty}

%%%%%%%%%%%%%%%%%%%%%%%%%%%%%%%%%%%%%%%%%%%%%%%%%%%%%%%%%%%%%%%%%%%%%%%%%%%%%%%%
\begin{abstract}

In reinforcement learning, reward shaping is an efficient way to guide the learning process of an agent, as the reward can indicate the optimal policy of the task. The potential-based reward shaping framework was proposed to guarantee policy invariance after reward shaping, where a potential function is used to calculate the shaping reward. In former work, we proposed a novel adaptive potential function (APF) method to learn the potential function concurrently with training the agent based on information collected by the agent during the training process, and examined the APF method in discrete action space scenarios. This paper investigates the feasibility of using APF in solving continuous-reaching tasks in a real-world robotic scenario with continuous action space. We combine the Deep Deterministic Policy Gradient (DDPG) algorithm and our proposed method to form a new algorithm called APF-DDPG. To compare APF-DDPG with DDPG, we designed a task where the agent learns to control Baxter's right arm to reach a goal position. The experimental results show that the APF-DDPG algorithm outperforms the DDPG algorithm on both learning speed and robustness.

\end{abstract}

%%%%%%%%%%%%%%%%%%%%%%%%%%%%%%%%%%%%%%%%%%%%%%%%%%%%%%%%%%%%%%%%%%%%%%%%%%%%%%%%
\section{INTRODUCTION}
\label{intro}

Reinforcement learning (RL) has come into force with the development of artificial intelligence in recent years. It is widely studied in various areas, such as gaming~\cite{Mnih2015, hafner2020dreamerv2, hessel2017rainbow}, robotics~\cite{openai2019solving, BaierLowenstein2007, millan2021robust}, and so on. 
RL aims to solve decision-making problems through trial and error. In particular, a critical component of RL is the reward function because it plays a vital role in deciding the policy quality that the agent can learn.
Therefore, in RL model building, the reward design is such a crucial part that popularizes the reward shaping method to incorporate domain knowledge into an RL agent so as to accelerate its learning process.

However, naive rewards, such as ad hoc rewards, may lead to unintended or even unsafe behaviors~\cite{randlov1998learning}. Therefore, the potential-based reward shaping (PBRS) framework was proposed to guarantee the policy invariance property where a potential function is introduced in the reward-shaping function~\cite{ng1999policy}.
This paper is fundamentally based on the PBRS framework.
Notably, we would like to mention the biological counterpart of states' potentials.
The valuation of the current state plays such a central role that animal brains contain a dedicated architecture around the amygdala~\cite{Pignatelli2019,MCLAUGHLIN2007}, using internal and perceptual state variables to anticipate approach vs. avoidance. Note that this concerns a computation in advance of the action, not solely an evaluation of a just-performed action.

In our former work \cite{Chen2021}, we proposed an adaptive potential function (APF) that learns the potential function concurrently with training an agent. The APF is learned based on the information collected during the training without external experts' help. 
We proved that APF can be well combined with two baseline algorithms: (1) Q-learning with multi-layer perceptrons (QMLP) and (2) Dueling deep Q-network (DQN). We combined the APF method with the two baselines to form two new RL algorithms: APF-QMLP and APF-Dueling-DQN. The experimental results showed that APF-QMLP and APF-Dueling-DQN outperform the QMLP and Dueling DQN in gaming environments with discrete action space, respectively.

\begin{figure}[t]
     \centering
     \includegraphics[width=.9\linewidth]{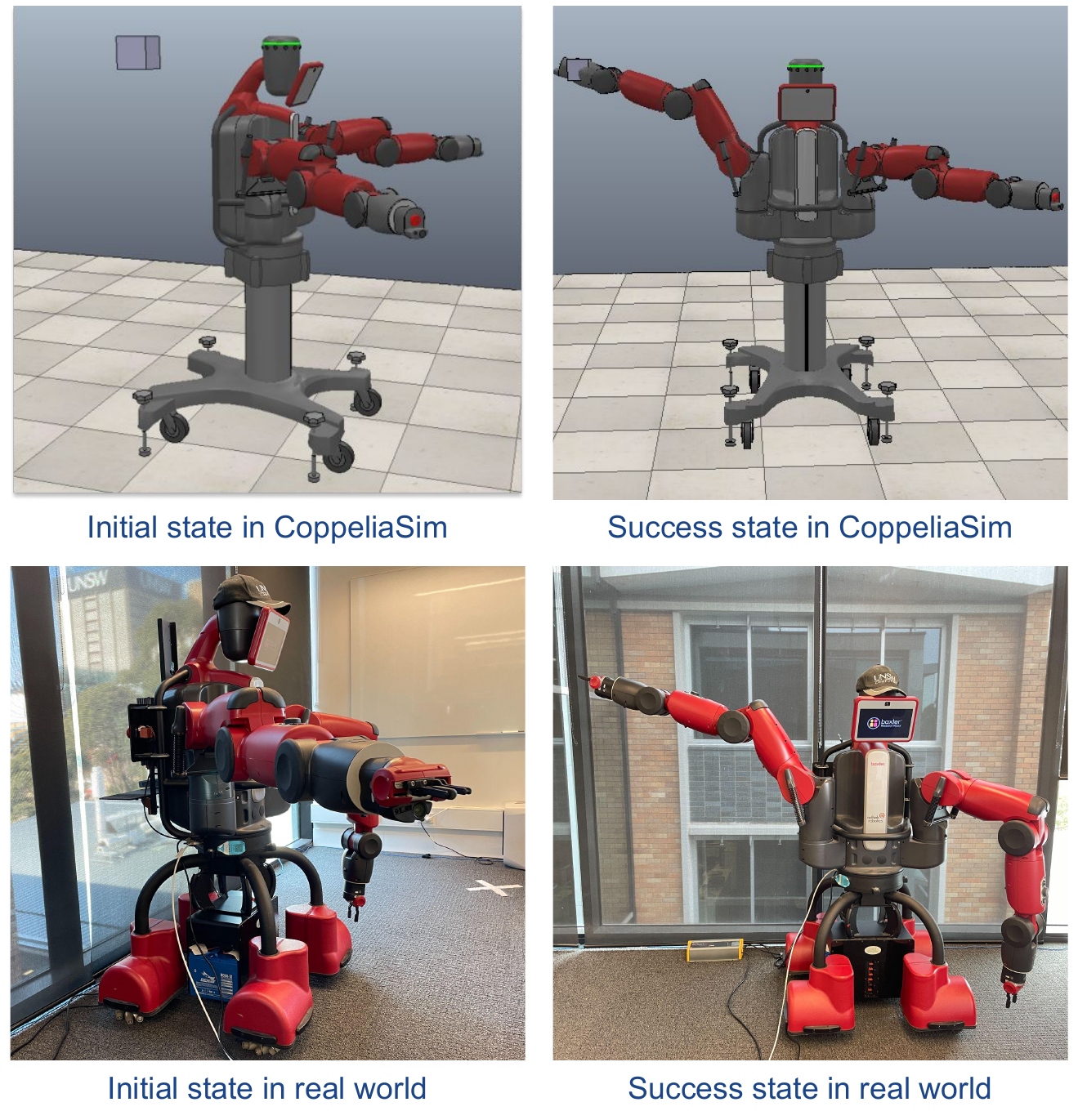}
     \caption{A visualization of the experimental environments with a Baxter robot in CoppeliaSim (first row) and the real world (second row). The task is to control Baxter's right arm's joints so that its right tip can reach the goal area, shown as a gray cube in the upper left corner of the first-row figures. The left-column figures show the initial state of Baxter, and the right-column figures give an example of a successful state in CoppeliaSim and the real world.}
     \label{fig:baxter_init}
\end{figure}

In this paper, we extend APF~\cite{Chen2021} to robotic environments, as shown in Fig. \ref{fig:baxter_init}.
Different from the tasks in \cite{Chen2021}, robotic environments usually have continuous state and action space.
Therefore, we combine the APF method with the Deep Deterministic Policy Gradient (DDPG) algorithm, designed for dealing with continuous action space tasks, to form the APF-DDPG algorithm. We examine the new APF-DDPG algorithm by solving a reaching task in the Baxter robot. The experiments are conducted in both a simulator and a real Baxter robot. Our results confirm the feasibility and efficiency of deploying the APF method in the continuous domain. Compared to the DDPG agent, the APF-DDPG agent achieves significantly higher performance and robustness. Implemented the APF-DDPG algorithm, the real robot acts smoothly and can quickly reach the goal, which shows that our method is robust in the real world.

{\bf Paper outline.} The remainder of this paper is organized as follows. First, section \ref{sec:relatedworks} summarizes the related works. Then, the background of RL, the PBRS theory, and the baseline DDPG are introduced in section \ref{sec:background}. The specific MDP setting in robotics is explained in section \ref{sec:mdp_robotics}. Section \ref{sec:methodology} describes the proposed APF-DDPG algorithm in detail, and section \ref{sec:experiments} shows the experimental results in both the simulator and the real world. At last, Section~\ref{sec:discussion} concludes this paper and discusses the potential problems and future research directions.

\section{Related Works}
\label{sec:relatedworks}

The theoretical basis of this paper is the PBRS framework proposed by Ng et al. in 1999. In \cite{ng1999policy}, the authors formalized the reward shaping function as a composition of potential functions and proved that the optimal policy is invariant under the PBRS framework.

The potential function has raked many researchers' attention. Combining human feedback with reward shaping was effectively carried out in~\cite{harutyunyan2015shaping} thanks to the PBRS framework. Moreover, reward shaping in multi-agent systems, e.g., \cite{babes2008social,devlin2011empirical}, also benefits from the PBRS framework,
as it reserves the Nash Equilibrium of multi-agent games~\cite{devlin2012dynamic}. Furthermore, Wiewiora et al. extended the potential function from the state domain to the state-action domain. They proved that initializing Q-values to a static potential function can achieve the same results as using it in reward shaping~\cite{wiewiora2003principled}.

In addition, Devlin and Kudenko showed that reward shaping with a dynamic potential function can still guarantee the policy invariance property. 
Then, the PBRS framework can be applied to more practical scenarios by learning a potential function dynamically rather than only using a static potential function. 

Reward shaping by learning potential values in abstract state space has been studied in several research works.
In \cite{grzes2008multigrid}, the authors proposed discretizing the continuous state space to an abstract state space and learning the value function of the abstract state space as a potential function.
However, this method can only compute the potential value for visited states. Similarly, Okudo and Yamada proposed a subgoal-based reward shaping method that abstracts the state space in a different way and learns the state values of this abstract state space as the potential values. In \cite{Okudo2023}, the abstract states were defined as the achievement status of specific subgoal states acquired from human experts.

Moreover, a few research works focus on deploying the PBRS framework in robotics. In~\cite{Jeon2023}, the authors benchmarked the standard form of PBRS in humanoid robots. Malysheva et al. trained a two-legged humanoid body to run using video-based shaping rewards. The agent successfully overcame sub-optimal running behaviors in human movements on videos by using the PBRS framework. Consequently, applying the PBRS framework in robotic RL is essential.

Inspired by these works, we adopt the idea of using an abstract state space to deploy APF in continuous spaces. However, instead of learning the value function as potentials, we propose to learn potential values by distinguishing good and bad past experiences in a robotic scenario.

\section{Background}
\label{sec:background}

This section first introduces the preliminaries of RL. Then, the PBRS framework is described.
%mathematically. 
Finally, we explain the RL algorithm DDPG, which is the baseline algorithm in this paper.

\subsection{Reinforcement Learning}
The standard reinforcement learning (RL) task is formalized using the Markov Decision Process (MDP), which is a five-element tuple $M = <S, A, R, \gamma, P>$. In an MDP, the environment is characterized by a set of states $S$, and the action space of the task is defined as $A$. $R: S \times A \times S \rightarrow \mathbb{R}$ is the environmental reward function that returns a real-value reward when the agent takes an action at the current state and transfers to the next state. The discount factor $\gamma$ is a real number within interval $[0,1]$, balancing the importance of short-term and long-term rewards. Finally, the transition probability $P(s'|s, a)$ gives the probability that the agent transfers to state $s'$ when it takes action $a$ at state $s$.

The goal of RL is to learn an optimal policy that maximizes the expected cumulative long-term reward. A policy $\pi: S \rightarrow \mathcal{P}(A)$ is a mapping from states to probability distributions over actions. 

Many frequently used RL algorithms are based on learning an optimal Q-function.
A Q-function is defined as in Equation \ref{eq:qfunc}, which indicates the future cumulative reward the agent can get when it takes an action $a$ at the state $s$. 

\begin{equation}\label{eq:qfunc}
    Q_\pi(s,a) \doteq \mathbb{E}_{a\sim \pi(s)}[\sum_{k=0}^{\infty} \gamma^k R_{t+k+1} \mid S_t=s, A_t=a]
\end{equation}

Then, we obtain the optimal Q-function $Q^{\ast}(s,a) = \sup_\pi Q_\pi(s,a)$,
%is obtained by updating the Q-function using the Bellman equation \cite{sutton2018book,Bellman1957}, 
where $\sup_\pi Q_\pi(s,a)$ is the supremum of Q-values in all policies.
Consequently, the optimal policy $\pi^{\ast}$ is obtained by selecting optimal actions using $\pi^{\ast}(s) = \arg \max_a Q^{\ast}(s,a)$.
% action that achieves the optimal Q-value, i.e., $Q_{\pi^\ast}(s,a)=Q^\ast (s,a)$.

\subsection{Potential-based Reward Shaping}
In RL, reward shaping is a common way to help accelerate the training process of an agent. To guarantee policy invariance after reward shaping, Ng et al. proposed the PBRS framework, where the crux is the reward shaping function $F: S \times S \rightarrow \mathbb{R}$ which is defined as follows:

\begin{equation}\label{eq:pbrs}
    F(s,s') = \gamma \Phi (s') - \Phi (s),
\end{equation}

\noindent where $\gamma$ is the same discount factor used in the MDP, and $\Phi: S\rightarrow \mathbb{R}$ is the potential function that takes a state and outputs the potential value of the state.
A potential function can be any function outputting a value for each state.
We aim to design a potential function that can describe how good each state is with respect to the task.

Equipped with the above reward shaping function, we build the new reward function as $R'(s,a,s') = R(s,a,s') +F(s,s')$. 
It is proved that the optimal policy in the original MDP $M = <S, A, R, \gamma, P>$ does not alter in the new MDP $M' = <S, A, R', \gamma, P>$ under the PBRS framework~\cite{ng1999policy}.

\subsection{Deep Deterministic Policy Gradient}

DDPG \cite{lillicrap2015continuous,silver2014deterministic} is an off-policy actor-critic algorithm that concurrently learns a Q-function (a.k.a. critic) and a policy (a.k.a. actor). There are four networks in the DDPG algorithm: (1) a Q-network $Q$ with parameter $\theta^Q$, (2) a target Q-network $\hat{Q}$ with parameter $\hat{\theta}^Q$, (3) a policy network $\mu$ with parameter $\theta^\mu$, and (4) a target policy network $\hat{\mu}$ with parameter $\hat{\theta}^\mu$. Then, the Q-network is updated by minimizing the loss function in Equation \ref{eq:loss_ddpg_q} by gradient descent.

\begin{equation}\label{eq:loss_ddpg_q}
    L_{\theta^Q} = \mathbb{E} [((Q(s,a \mid \theta^Q) - (r + \gamma \hat{Q}(s', \hat{\mu}(s' \mid \hat{\theta}^\mu) \mid \hat{\theta}^Q))^2]
\end{equation}

The policy network $\mu$ is updated by solving Equation \ref{eq:ddpg_mu} with gradient ascent.

\begin{equation}\label{eq:ddpg_mu}
    \max_{\theta^\mu} \mathbb{E} [Q(s, \mu(s \mid \theta^\mu) \mid \theta^Q)]
\end{equation}
We employ DDPG in this robotic scenario because it is especially designed for environments with continuous action space.

\section{MDP in the robotic scenario}
\label{sec:mdp_robotics}

\begin{figure}[t]
     \centering
     \includegraphics[width=.9\linewidth]{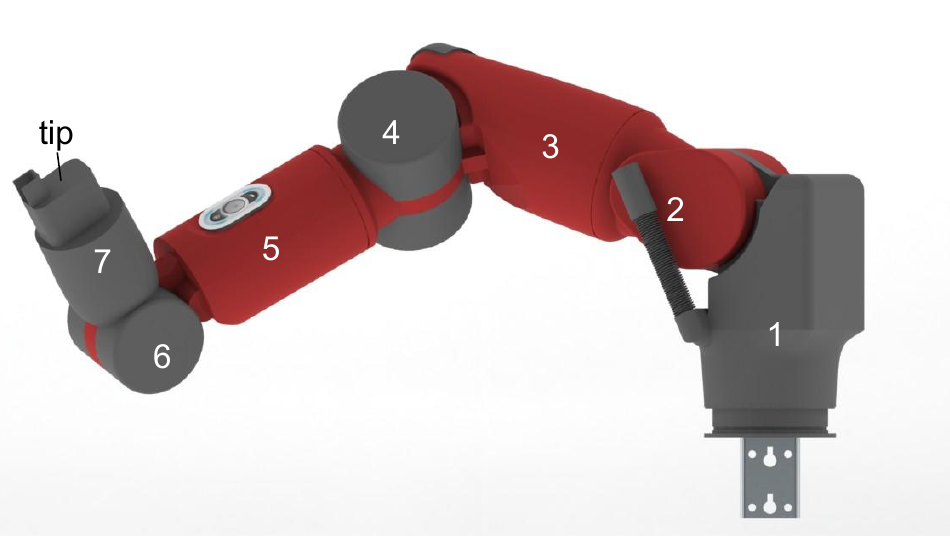}
     \caption{An image of Baxter's right arm. Each joint is sequentially numbered from $1$ to $7$ from shoulder to wrist. The tip is at the end of the arm.}
     \label{fig:r_arm}
\end{figure}

In this section, we introduce how we apply our RL algorithm on the Baxter robot (as in Fig. \ref{fig:baxter_init}) in our experiment. Baxter is a two-armed robot with seven Degrees of Freedom (DoF) on each arm manipulator. We sequentially number each joint from shoulder to wrist as illustrated in Fig. \ref{fig:r_arm}. In this experiment, we control the first, the second, and the fourth joint of Baxter's right arm in a continuous space.

To define the MDP of our experiment, we set the discount factor to $\gamma = 0.99$, and the environment to deterministic, i.e., the transition probability $P(s'\mid s,a)$ is degenerated to a single-point distribution.
In other words, the agent deterministically transfers to state $s'$ when taking action $a$ at state $s$.
%and the discount factor to $\gamma = 0.99$.
The state space $S$ and action space $A$ in this experiment are both continuous.  An environmental state $s\in S$ is represented as a six-dimensional vector as shown in Equation \ref{eq:state}, where the first three elements $x_{tip}, y_{tip}, z_{tip}$ denote the position of the robot's right tip in the three-dimensional Euclidean space, and $j_1, j_2$ and $j_4$ denote the angular position in radians of the first, the second and the fourth joints, respectively.
The origin of the Euclidean coordinate is the position of the first joint.
%system is set in the same position as the first joint and the same orientation as the first joint at $0.0$ radians.

\begin{equation}\label{eq:state}
s = (x_{tip}, y_{tip}, z_{tip}, j_1, j_2, j_4)
\end{equation}

Given a state, the robot may take an action which can be represented as a three-dimensional vector as shown in Equation \ref{eq:action}, where $\delta_1, \delta_2$ and $\delta_4$ denote the radians the corresponding joints will change, respectively. To enable robotic movements smooth and natural, each joint is only allowed to change within an interval of $[-\frac{\pi}{16},\frac{\pi}{16}]$ radians at each time step.

\begin{equation}\label{eq:action}
a = (\delta_1, \delta_2, \delta_4),\ \delta_1, \delta_2, \delta_4  \in [-\frac{\pi}{16},\frac{\pi}{16}]
\end{equation}

The environmental reward function is defined as Equation \ref{eq:reward}, where $Dis(s,s_g)$ is defined as the Euclidean distance between the robot's right tip at the current state $s$ and the goal position.
Specifically, the agent gets a reward of $1$ when it reaches the goal state $s_g$ with a tolerance of $0.1$ meters. If a collision happens, the agent receives a reward of $-max\_steps$ and the current training episode terminates, where $max\_steps$ is $100$ in the experiment. 
%Define $Dis(s,s_g)$ as the distance between the robot's right tip and the goal position. 
When $Dis(s,s_g)$ is farther than $1.5$ meters, the agent gets $-5$ for each step. For $Dis(s,s_g)$ within $[0.1,0.5)$ meters, the agent acquires $-0.5$ for each step. For other circumstances where $Dis(s,s_g)$ is between $[0.5,1.5]$ meters, the agent gets $-1$ for each step.

\begin{equation}\label{eq:reward}
    R(s) = \left\{\begin{matrix}
           1 & Dis(s,s_g)<0.1\\ 
           -0.5 & 0.1 \leq Dis(s,s_g)<0.5\\ 
           -1 & 0.5 \leq Dis(s,s_g) \leq 1.5\\
           -5 & Dis(s,s_g)>1.5\\ 
           -max\_steps & \text{collision}
       \end{matrix}\right.
\end{equation}

\begin{figure*}[t]
     \centering
     \includegraphics[width=.9\linewidth]{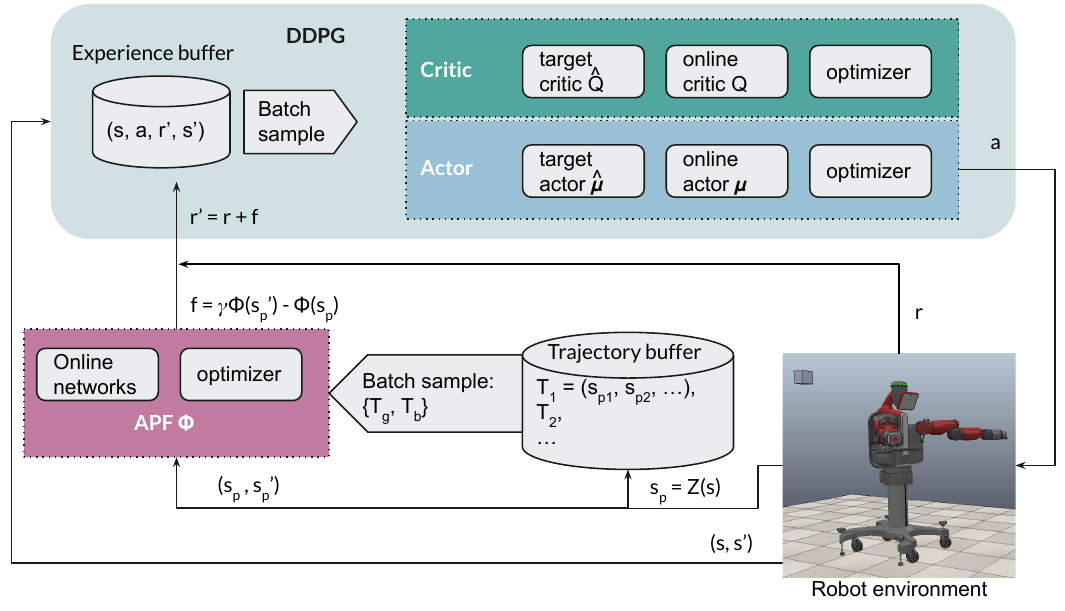}
     \caption{A schematic of APF-DDPG. The APF network is trained to output a potential value for each state. Then, the environmental reward is shaped based on the PBRS framework, and the shaped reward is collected to train the underlying DDPG networks. After training, the actor network can be used to control the Baxter to reach the goal in both the simulator and the real world.}
     \label{fig:apf_ddpg}
\end{figure*}

\section{Methodology}
\label{sec:methodology}

In previous work \cite{Chen2021}, our proposed adaptive potential function (APF) learned the potential function based on the information collected by the agent itself during training. We evaluated APF by solving problems in discrete action space. This paper extends APF in solving a robotic task in a continuous action space.

\subsection{Discrete Potential States}

To extend APF in continuous state space, we define a state mapping function $Z: S \rightarrow S_p$ to project the continuous six-dimensional state space $S$ to a discrete three-dimensional state space $S_p$, where an environmental state $s \in S$ is defined in Equation \ref{eq:state}. Each mapped state $s_p \in S_p$ is defined as the rounded position of the robot's right tip $s_p = (\lfloor x_{tip} \rfloor, \lfloor y_{tip} \rfloor, \lfloor z_{tip} \rfloor)$, where the {\em floor} operator $\lfloor \cdot \rfloor$ rounds Baxter's tip position to a discrete space by scale of $0.1$ meter. We call such states $s_p$ {\em potential states} to distinguish them from the environmental states. Note that the APF is learned in the three-dimensional discrete potential state space, while the underlying RL algorithm is learned in the six-dimensional continuous environmental state space.

We introduce the state mapping function $Z$ to simplify the representation of states from six-dimensional vectors to three-dimensional vectors for two reasons: (\textit{i}) Learning APF needs the operation of counting states, but counting continuous states is not meaningful as most of them will only occur once. (\textit{ii}) As the goal state does not require specific angles of the controlled joints, discrete position information of the right tip is enough for building a potential space to guide the agent in a high-level direction.

\subsection{Adaptive Potential Function}

The APF $\Phi: S \rightarrow \mathbb{R}$ takes a state as input and outputs the potential value of that state. To generate the potential value computation method on unvisited states, we define a neural network with parameter $\phi$ to approximate the APF $\Phi$. The updating process of the APF network in this experiment is as follows.

During the $i$-th episode of the training, the agent records the sequence of all potential states visited in this episode in a trajectory $t_i = (s_{p1}, s_{p2}, ...)$. We store trajectories in a priority queue, where the trajectories are ordered by the corresponding episodic reward $R_{ei}$, i.e., the cumulative reward of the episode, of each trajectory $t_i$. 
Ordering from the highest to the lowest episodic rewards, the first $50\%$ trajectories are defined as good trajectories, and the other $50\%$ are bad trajectories. 

After each episode, a set of half-batch-size good trajectories (denoted as $T_g$) and a set of half-batch-size bad trajectories (denoted as $T_b)$ are randomly sampled. For each state $s_p$ appeared in $T_g$ and $T_b$, we define its occurrence in $T_g$ as $N_g[s_p]$ and its occurrence in $T_b$ as $N_b[s_p]$, respectively. Note that a state can appear in both good and bad trajectories. Therefore, the APF network is updated by minimizing the loss function in Equation \ref{eq:loss_pf}.

\begin{equation}\label{eq:loss_pf}
L_{\phi} = \mathbb{E}\left[\left(\Phi(s_p; \phi) - \frac{N_{g}[s_p] - N_{b}[s_p]}{N_{g}[s_p] + N_{b}[s_p]}\right)^2\right],
\end{equation}

\subsection{APF-DDPG}

This section introduces our new algorithm, APF-DDPG, which incorporates the abovementioned APF into the classic DDPG algorithm.
We depict the APF-DDPG algorithm in Fig. \ref{fig:apf_ddpg} for the most straightforward illustration.
The figure shows that the APF-DDPG agent learns the APF network concurrently with the DDPG networks. In specific, the APF network $\Phi$ is updated every episode, while the four DDPG networks, i.e., the critic network, the target critic network, the actor network, and the target actor network, are updated every time step.

%We apply the proposed APF in the classic DDPG algorithm to form a new algorithm termed APF-DDPG.
%As shown in Fig. \ref{fig:apf_ddpg}, the APF-DDPG agent learns the APF network concurrently with the DDPG networks. In specific, the APF network $\Phi$ is updated every episode, and the DDPG networks: the critic network, the target critic network, the actor network, and the target actor network, are updated every time step.

When the APF-DDPG agent takes an action $a$ at the state $s$ and is transferred to the next state $s'$, the environment returns a reward $r$ while the APF network returns a shaping reward $f = \Phi(Z(s')) - \Phi(Z(s))$. Then, the APF-DDPG agent collects the shaped reward $r' = r + f$ into the experience buffer to update DDPG networks and collects the mapped potential state $s_p=Z(s)$ into the trajectory buffer to update the APF network.

This loop repeats until the end of training. The role of APF in this loop is to extract information from the agent's past experiences to boost the learning process of the DDPG networks. When the training finishes, the trained actor network can be applied to control Baxter in both the simulator and the real world.

\section{Experiments}
\label{sec:experiments}

We design a reaching task to compare the APF-DDPG agent and the DDPG agent using the Baxter robot. The RL agents are trained in the CoppeliaSim simulation platform (see the first row in Fig. \ref{fig:baxter_init}) on an Ubuntu 22.04 system using a Nvidia 2080 Ti graphics card. Then, we also examine the trained RL models on a real Baxter robot, as shown in the second row of Fig. \ref{fig:baxter_init}.

%As shown in the first row of Fig. \ref{fig:baxter_init},
Our task is to navigate Baxter's right arm so that its tip can reach the goal area as soon as possible. The goal area is visualized as the gray cube in the upper left corner of the first-row figures in Fig. \ref{fig:baxter_init}.

\subsection{Experimental Parameters}

Both the DDPG agent and the APF-DDPG agent are repeatedly trained for $20$ experimental runs. In each experimental run, the agent is trained for $2000$ episodes with a maximal $100$ steps ($max\_steps = 100$) in each episode.

In both DDPG and APF-DDPG, the critic network and the actor network are implemented using multi-layer perceptrons (MLPs). The critic network takes a state-action pair as input and outputs a Q-value, while the actor network takes a state as input and generates an action for the agent to execute. Both the critic network and the actor network have two hidden layers, each of which has $512$ neurons and uses ReLU as the activation function. Additionally, APF-DDPG uses an extra MLP to implement the APF network. The MLP of APF has two hidden layers with neuron sizes of $512$ and $256$, and uses ReLU as the activation function for each layer. 

Furthermore, the learning rate for both the critic network and the APF network is $0.02$, and that for the actor network is $0.01$. The experience replay buffer stores the most recent $10000$ experiences, while the trajectory buffer stores the $2000$ best trajectories. The batch size for updating all the networks is $64$.

\begin{figure}[t]
     \centering
     \includegraphics[width=.9\linewidth]{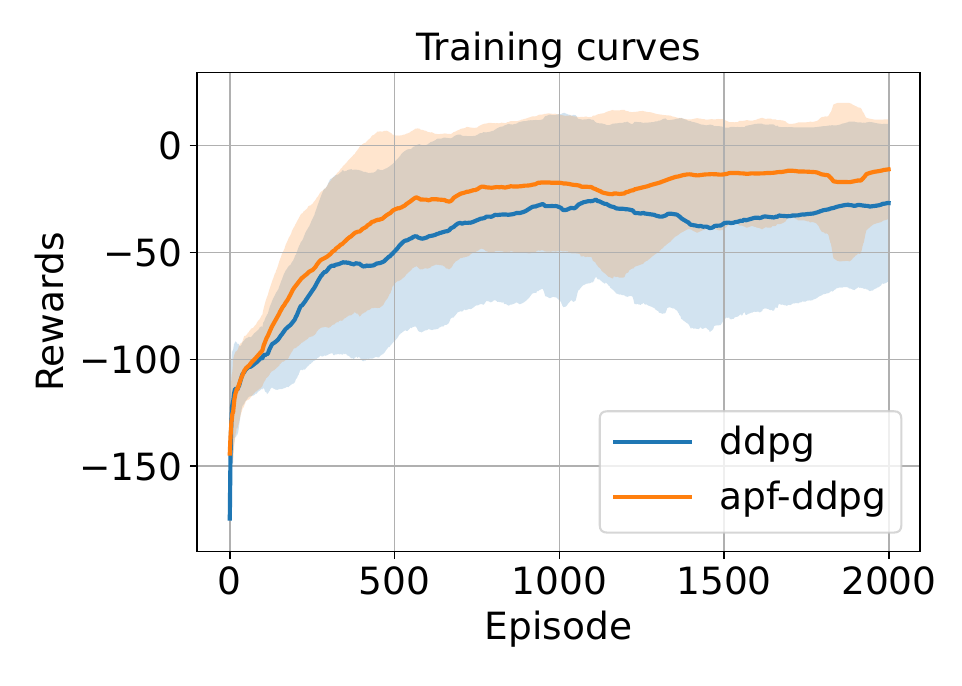}
     \caption{Comparison of performances of the DDPG agent (the blue curve) and the APF-DDPG agent (the orange curve). Each curve is averaged over 20 experimental runs and each run is smoothed by an average window of 100 episodes. The shaded region represents the standard deviation range. }
     \label{fig:res}
\end{figure}

\subsection{Experimental Results}

We compare the performances of the APF-DDPG agent with the DDPG agent using the MDP settings introduced in Section \ref{sec:mdp_robotics}. Fig. \ref{fig:res} shows the average episodic rewards over 20 runs for each agent. The blue curve represents the average performance of the DDPG agent, and the orange curve represents the average performance of the APF-DDPG agent. The standard deviation range is visualized as the shaded region. It can be observed that the APF-DDPG agent outperforms the DDPG agent.

\begin{figure}[b]
     \centering
     \includegraphics[width=.9\linewidth]{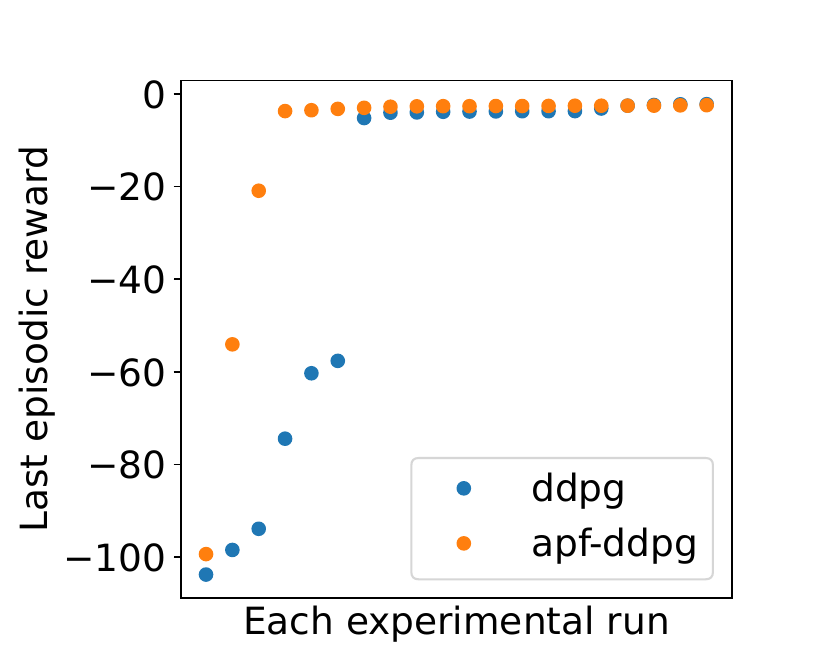}
     \caption{Comparison among the averaged cumulative reward over the last $100$ episodes in each of the 20 experiments for the DDPG agent and the APF-DDPG agent. Results are shown in ascending order for each agent. Blue dots represent DDPG, while orange dots represent APF-DDPG.}
     \label{fig:last_r}
\end{figure}
\begin{figure*}[t]
     \centering
     \includegraphics[width=.9\linewidth]{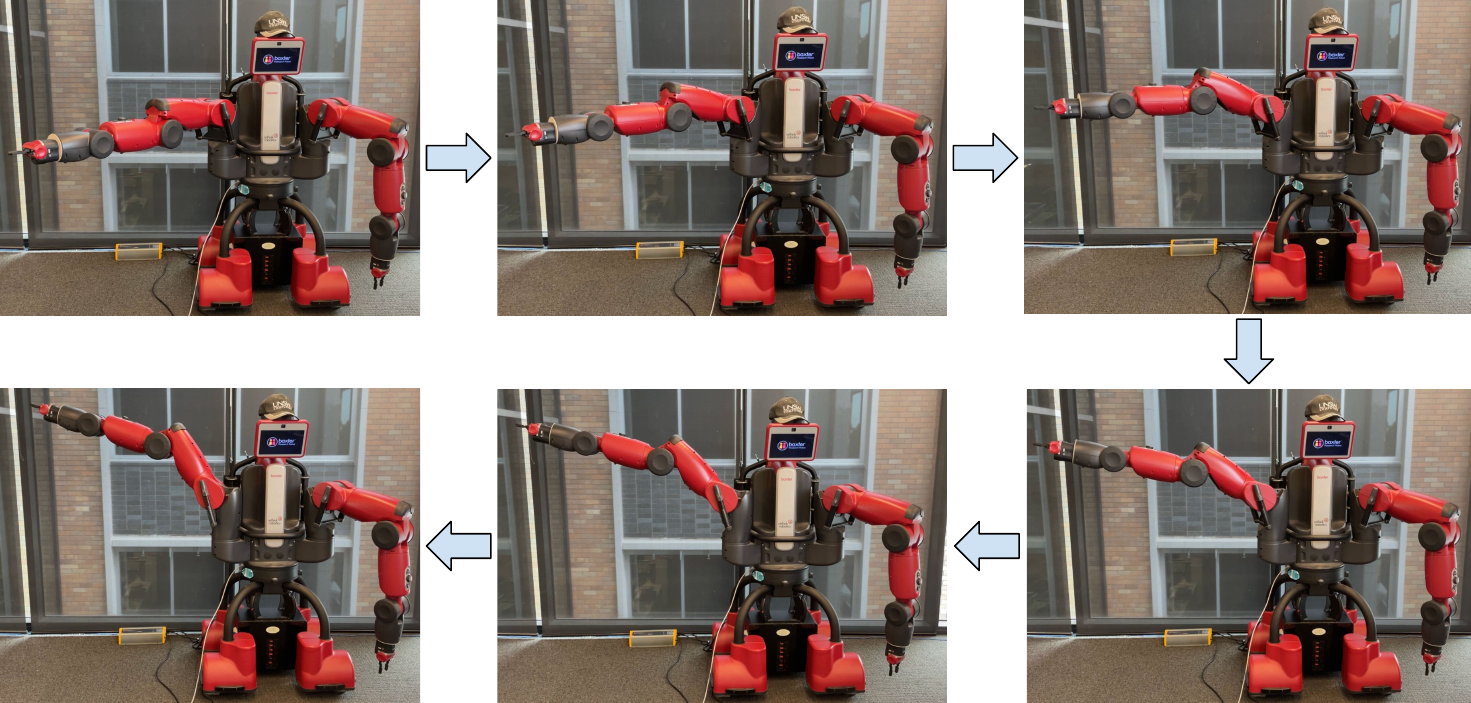}
     \caption{Steps employed by the real Baxter robot to reach the goal area using the trained APF-DDPG model.}
     \label{fig:real}
\end{figure*}

Additionally, we perform the Student's t-test on the averaged data of the 20 DDPG runs and the averaged data of the 20 APF-DDPG runs. The t-statistic is $-19.1$, and the p-value is $5.8e-78$, which confirms the significance of the difference between the performances of the DDPG agent and the APF-DDPG agent.

Furthermore, Fig. \ref{fig:last_r} shows the averaged cumulative reward over the last $100$ episodes of each experimental run of each agent. The results are plotted in ascending order for a clear comparison. The DDPG results are represented in blue dots, and the APF-DDPG results are shown in orange.
We can observe that, overall, the APF-DDPG agent performs better. For well-performed runs, both DDPG and APF-DDPG reach the optimal cumulative reward. However, the DDPG agent is less stable in the sense that it fails more frequently than the APF-DDPG agent. Specifically, the DDPG agent fails six out of twenty runs, while the APF-DDPG agent only fails two out of twenty runs.

\subsection{Experiments on A Real Baxter}

To further test the robustness of the APF-DDPG agent, we implement it in a real Baxter. Because the Baxter robot’s software development kit (SDK) is old and stopped updating since 2014, we control the robot using the Robot Operating System (ROS) Indigo on a Ubuntu 14.04 system.

In this practical implementation, we utilize the trained actor network to generate an action based on each state, where the state and action settings are kept the same as in the CoppeliaSim simulation. Implementing an algorithm in a real robot differs from that in a simulator, from which all information about the robot and the environment is immediately accessible.
Particularly, because there is no sensor to get the tip position in the real Baxter's experiment, we calculate it based on the seven joints' angles using forward kinematics. 
Besides, we must ensure that each action is appropriate so that the robot would not act beyond its limitation to cause any damage or danger.

As illustrated in Fig. \ref{fig:real}, the real Baxter finishes the reaching task in five steps as expected\footnote{Please see our complementary video for more details of the experiment:~\url{https://drive.google.com/file/d/1HeWukevcMCQPBaVhBNttmX0O6iqPD94t/view?usp=sharing}}.
During the approaching, the robot's right arm smoothly moved from the initial position until its tip reached the goal area without abundant action.
This shows our model can be robustly applied in practical scenarios.

\section{Conclusion and future works}
\label{sec:discussion}

This paper proves that the APF proposed in~\cite{Chen2021} can accelerate the RL algorithms in continuous action space scenarios. 
By introducing a discrete potential state space, we successfully combine the APF with the DDPG algorithm to form a new RL algorithm, APF-DDPG, which can deal with problems in continuous action space. 

To compare the APF-DDPG agent with the DDPG agent, we designed a robotic reaching task in a Baxter robot. The RL agents are trained in the CoppeliaSim simulation and further tested in a real Baxter robot.
The results show that the APF-DDPG agent not only outperforms the DDPG agent but also manages to achieve the goal more frequently.

As for the future directions extending this work, we would like to investigate the possibility of enhancing the exploration ability of APF-based agents.
Consider that the agent learns and uses the APF simultaneously, and the APF is learned from a limited number of best-performing trajectories.
Utilizing the APF too early might lead to the agent being trapped in local optima.
In this paper, the epsilon-greedy strategy is sufficient to keep the agent out of the trap.
However, we cannot guarantee this in all tasks.
In the future, we would like to combine the algorithm with more powerful exploration methods to assist the APF-based agent to avoid being trapped in local optima.



\section*{ACKNOWLEDGMENT}

Yifei Chen credits Marco Wiering's role in polishing the idea of adaptive potential function. Yifei Chen thanks Yuzhe Zhang for improving the writing of this paper.

%%%%%%%%%%%%%%%%%%%%%%%%%%%%%%%%%%%%%%%%%%%%%%%%%%%%%%%%%%%%%%%%%%%%%%%%%%%%%%%%

\bibliographystyle{IEEEtran} 
\bibliography{IEEEabrv,mybibfile}

\begin{thebibliography}{10}
\providecommand{\url}[1]{#1}
\csname url@rmstyle\endcsname
\providecommand{\newblock}{\relax}
\providecommand{\bibinfo}[2]{#2}
\providecommand\BIBentrySTDinterwordspacing{\spaceskip=0pt\relax}
\providecommand\BIBentryALTinterwordstretchfactor{4}
\providecommand\BIBentryALTinterwordspacing{\spaceskip=\fontdimen2\font plus
\BIBentryALTinterwordstretchfactor\fontdimen3\font minus \fontdimen4\font\relax}
\providecommand\BIBforeignlanguage[2]{{%
\expandafter\ifx\csname l@#1\endcsname\relax
\typeout{** WARNING: IEEEtran.bst: No hyphenation pattern has been}%
\typeout{** loaded for the language `#1'. Using the pattern for}%
\typeout{** the default language instead.}%
\else
\language=\csname l@#1\endcsname
\fi
#2}}

\bibitem{Mnih2015}
\BIBentryALTinterwordspacing
V.~Mnih, K.~Kavukcuoglu, D.~Silver, A.~A. Rusu, J.~Veness, M.~G. Bellemare, A.~Graves, M.~Riedmiller, A.~K. Fidjeland, G.~Ostrovski, S.~Petersen, C.~Beattie, A.~Sadik, I.~Antonoglou, H.~King, D.~Kumaran, D.~Wierstra, S.~Legg, and D.~Hassabis, ``Human-level control through deep reinforcement learning,'' \emph{Nature}, vol. 518, no. 7540, pp. 529--533, Feb. 2015. [Online]. Available: \url{https://doi.org/10.1038/nature14236}
\BIBentrySTDinterwordspacing

\bibitem{hafner2020dreamerv2}
D.~Hafner, T.~Lillicrap, M.~Norouzi, and J.~Ba, ``Mastering atari with discrete world models,'' \emph{arXiv preprint arXiv:2010.02193}, 2020.

\bibitem{hessel2017rainbow}
\BIBentryALTinterwordspacing
M.~Hessel, J.~Modayil, H.~van Hasselt, T.~Schaul, G.~Ostrovski, W.~Dabney, D.~Horgan, B.~Piot, M.~Azar, and D.~Silver, ``Rainbow: Combining improvements in deep reinforcement learning,'' 2017. [Online]. Available: \url{https://arxiv.org/abs/1710.02298}
\BIBentrySTDinterwordspacing

\bibitem{openai2019solving}
OpenAI, I.~Akkaya, M.~Andrychowicz, M.~Chociej, M.~Litwin, B.~McGrew, A.~Petron, A.~Paino, M.~Plappert, G.~Powell, R.~Ribas, J.~Schneider, N.~Tezak, J.~Tworek, P.~Welinder, L.~Weng, Q.~Yuan, W.~Zaremba, and L.~Zhang, ``Solving rubik's cube with a robot hand,'' 2019.

\bibitem{BaierLowenstein2007}
T.~Baier-Lowenstein and J.~Zhang, ``Learning to grasp everyday objects using reinforcement-learning with automatic value cut-off,'' in \emph{2007 {IEEE}/{RSJ} International Conference on Intelligent Robots and Systems}.\hskip 1em plus 0.5em minus 0.4em\relax {IEEE}, Oct. 2007.

\bibitem{millan2021robust}
C.~C. Millan-Arias, B.~J. Fernandes, F.~Cruz, R.~Dazeley, and S.~Fernandes, ``A robust approach for continuous interactive actor-critic algorithms,'' \emph{IEEE Access}, vol.~9, pp. 104\,242--104\,260, 2021.

\bibitem{randlov1998learning}
J.~Randl{\o}v and P.~Alstr{\o}m, ``Learning to drive a bicycle using reinforcement learning and shaping.'' in \emph{ICML}, vol.~98, 1998, pp. 463--471.

\bibitem{ng1999policy}
A.~Y. Ng, D.~Harada, and S.~Russell, ``Policy invariance under reward transformations: Theory and application to reward shaping,'' in \emph{International Conference on Machine Learning (ICML)}, vol.~99, 1999, pp. 278--287.

\bibitem{Pignatelli2019}
\BIBentryALTinterwordspacing
M.~Pignatelli and A.~Beyeler, ``Valence coding in amygdala circuits,'' \emph{Current Opinion in Behavioral Sciences}, vol.~26, pp. 97--106, Apr. 2019. [Online]. Available: \url{https://doi.org/10.1016/j.cobeha.2018.10.010}
\BIBentrySTDinterwordspacing

\bibitem{MCLAUGHLIN2007}
\BIBentryALTinterwordspacing
R.~McLaughlin and S.~Floresco, ``The role of different subregions of the basolateral amygdala in cue-induced reinstatement and extinction of food-seeking behavior,'' \emph{Neuroscience}, vol. 146, no.~4, pp. 1484--1494, 2007. [Online]. Available: \url{https://www.sciencedirect.com/science/article/pii/S0306452207003454}
\BIBentrySTDinterwordspacing

\bibitem{Chen2021}
\BIBentryALTinterwordspacing
Y.~Chen, H.~Kasaei, L.~Schomaker, and M.~Wiering, ``Reinforcement learning with potential functions trained to discriminate good and bad states,'' in \emph{2021 International Joint Conference on Neural Networks ({IJCNN})}.\hskip 1em plus 0.5em minus 0.4em\relax {IEEE}, July 2021. [Online]. Available: \url{https://doi.org/10.1109/ijcnn52387.2021.9533682}
\BIBentrySTDinterwordspacing

\bibitem{harutyunyan2015shaping}
A.~Harutyunyan, T.~Brys, P.~Vrancx, and A.~Now{\'e}, ``Shaping mario with human advice,'' in \emph{Proceedings of the 2015 International Conference on Autonomous Agents and Multiagent Systems}.\hskip 1em plus 0.5em minus 0.4em\relax Citeseer, 2015, pp. 1913--1914.

\bibitem{babes2008social}
M.~Babes, E.~M. de~Cote, and M.~L. Littman, ``Social reward shaping in the prisoner's dilemma,'' in \emph{Proceedings of the International Joint Conference on Autonomous Agents and Multi Agent Systems (AAMAS)}, 2008.

\bibitem{devlin2011empirical}
S.~Devlin, D.~Kudenko, and M.~Grze{\'s}, ``An empirical study of potential-based reward shaping and advice in complex, multi-agent systems,'' \emph{Advances in Complex Systems}, vol.~14, no.~02, pp. 251--278, 2011.

\bibitem{devlin2012dynamic}
S.~M. Devlin and D.~Kudenko, ``Dynamic potential-based reward shaping,'' in \emph{Proceedings of the 11th International Conference on Autonomous Agents and Multiagent Systems}.\hskip 1em plus 0.5em minus 0.4em\relax IFAAMAS, 2012, pp. 433--440.

\bibitem{wiewiora2003principled}
E.~Wiewiora, G.~W. Cottrell, and C.~Elkan, ``Principled methods for advising reinforcement learning agents,'' in \emph{Proceedings of the 20th International Conference on Machine Learning (ICML-03)}, 2003, pp. 792--799.

\bibitem{grzes2008multigrid}
M.~Grze{\'s} and D.~Kudenko, ``Multigrid reinforcement learning with reward shaping,'' in \emph{International Conference on Artificial Neural Networks (ICANN)}.\hskip 1em plus 0.5em minus 0.4em\relax Springer, 2008, pp. 357--366.

\bibitem{Okudo2023}
\BIBentryALTinterwordspacing
T.~Okudo and S.~Yamada, ``Learning potential in subgoal-based reward shaping,'' \emph{{IEEE} Access}, vol.~11, pp. 17\,116--17\,137, 2023. [Online]. Available: \url{https://doi.org/10.1109/access.2023.3246267}
\BIBentrySTDinterwordspacing

\bibitem{Jeon2023}
\BIBentryALTinterwordspacing
S.~H. Jeon, S.~Heim, C.~Khazoom, and S.~Kim, ``Benchmarking potential based rewards for learning humanoid locomotion,'' in \emph{2023 {IEEE} International Conference on Robotics and Automation ({ICRA})}.\hskip 1em plus 0.5em minus 0.4em\relax {IEEE}, May 2023. [Online]. Available: \url{https://doi.org/10.1109/icra48891.2023.10160885}
\BIBentrySTDinterwordspacing

\bibitem{lillicrap2015continuous}
T.~P. Lillicrap, J.~J. Hunt, A.~Pritzel, N.~Heess, T.~Erez, Y.~Tassa, D.~Silver, and D.~Wierstra, ``Continuous control with deep reinforcement learning,'' in \emph{International Conference on Learning Representations (ICLR)}, 2016.

\bibitem{silver2014deterministic}
D.~Silver, G.~Lever, N.~Heess, T.~Degris, D.~Wierstra, and M.~Riedmiller, ``Deterministic policy gradient algorithms,'' in \emph{International Conference on Machine Learning (ICML)}.\hskip 1em plus 0.5em minus 0.4em\relax Pmlr, 2014, pp. 387--395.

\end{thebibliography}

\addtolength{\textheight}{-12cm}   % This command serves to balance the column lengths

\end{document}